\documentclass[10pt,twocolumn,letterpaper]{article}

\usepackage{cvpr}              % To produce the CAMERA-READY version
\definecolor{cvprblue}{rgb}{0.21,0.49,0.74}
\usepackage[pagebackref,breaklinks,colorlinks,allcolors=cvprblue]{hyperref}
\usepackage{booktabs}
\usepackage[table]{xcolor}
\usepackage{multirow, pifont, enumitem}
\usepackage[ruled, vlined, noline, noend]{algorithm2e}
\def\paperID{} % *** Enter the Paper ID here
\def\confName{CVPR}
\def\confYear{2026}

\title{Patch as Node: Human-Centric Graph Representation Learning \\for Multimodal Action Recognition}

\author{Zeyu Liang \quad Hailun Xia\footnotemark \quad Naichuan Zheng \\
 School of Information and Communication Engineering\\
Beijing University of Posts and Telecommunications\\
{\tt\small \{lzy\_sfading, xiahailun, 2022110134zhengnaichuan\}@bupt.edu.cn}}

\begin{document}
\maketitle
\begin{abstract}
\indent While human action recognition has witnessed notable achievements, multimodal methods fusing RGB and skeleton modalities still suffer from their inherent heterogeneity and fail to fully exploit the complementary potential between them. In this paper, we propose \textbf{PAN}, the first human-centric graph representation learning framework for multimodal action recognition, in which token embeddings of RGB patches containing human joints are represented as spatiotemporal graphs. The human-centric graph modeling paradigm suppresses the redundancy in RGB frames and aligns well with skeleton-based methods, thus enabling a more effective and semantically coherent fusion of multimodal features. Since the sampling of token embeddings heavily relies on 2D skeletal data, we further propose attention-based post calibration to reduce the dependency on high-quality skeletal data at a minimal cost in terms of model performance. To explore the potential of PAN in integrating with skeleton-based methods, we present two variants: \textbf{PAN-Ensemble}, which employs dual-path graph convolution networks followed by late fusion, and \textbf{PAN-Unified}, which performs unified graph representation learning within a single network. On three widely used multimodal action recognition datasets, both PAN-Ensemble and PAN-Unified achieve state-of-the-art (SOTA) performance in their respective settings of multimodal fusion: separate and unified modeling, respectively.

\footnotetext{* \quad Corresponding author.}
% 结尾加一句意义和展望 Skeleton data is naturally human-centric and invariant to viewpoint changes, yet it lacks the rich appearance information that can be captured from RGB data.
% and raw 3D skeleton coordinates 
\end{abstract}

% \indent While human action recognition has witnessed notable achievements, multimodal methods fusing RGB and skeleton modalities still suffer from their inherent heterogeneity and fail to fully exploit the complementary potential between them. In this paper, we propose \textbf{PAN}, the first human-centric graph representation learning framework for video action recognition, in which token embeddings of RGB patches containing human joints are represented as spatiotemporal graphs. The human-centric graph modeling paradigm suppresses the redundancy in RGB frames and aligns well with GCN-based methods in skeleton-based action recognition, thus facilitating more effective and semantically coherent fusion of multimodal features. Since the sampling of token embeddings heavily relies on 2D skeletal data, we further propose attention-based post calibration to eliminate the dependency on high-quality data at a minimal cost in terms of model performance. To exploit the potential of PAN in multimodal action recognition, we present two variants: \textbf{PAN-Ensemble}, which employs dual-path graph convolution networks followed by late fusion, and \textbf{PAN-Unified}, which performs unified graph representation learning within a single network. On three widely used multimodal action recognition datasets, both PAN-Ensemble and PAN-Unified achieve state-of-the-art (SOTA) performance in their respective settings of multimodal fusion: late fusion and unified modeling, respectively. The code will be available soon.    
\section{Introduction}
\label{sec:intro}

\begin{figure}[t]
\centering
\includegraphics[width=1.0\linewidth]{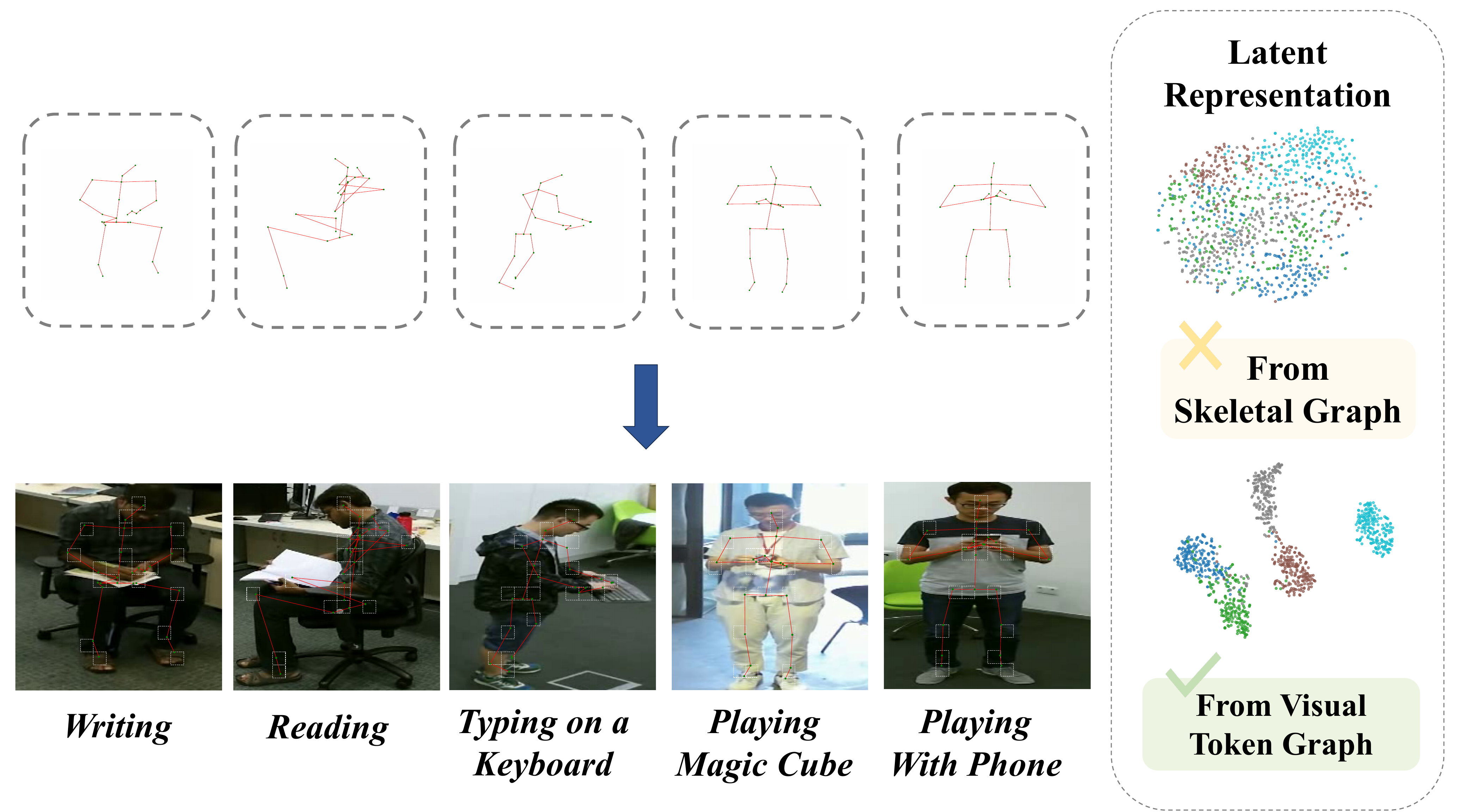}
\caption{\textbf{PAN constructs visual token graphs based on 2D skeletal data.} The learned representations can effectively distinguish five actions which are challenging for skeletal graphs, even when viewpoint changes lead to visually similar appearances.}
\label{fig:intro}
\end{figure}

Human action recognition is a fundamental task in computer vision, with a wide range of applications such as intelligent monitoring \cite{elkholy2019efficient}, human-computer interaction \cite{liu2017human}, and virtual reality \cite{bates2017line}. In the past few years, the integration of RGB and skeleton modalities has received a lot of attention \cite{ahn2023star,das2020vpn,duan2022revisiting,reilly2024just} due to their complementary potential. 

Specifically, the intrinsic human-centric nature of skeletal data provides a succinct and robust representation of human behaviors, resilient to environmental changes such as lighting variations and viewpoint changes. Yet, skeleton-based methods suffer from the absence of textual and appearance information and are highly dependent on data quality. In contrast, RGB frames contain rich appearance cues but are often redundant and vulnerable to environmental changes, forming a natural complement to skeletal data.

% as illustrated in Fig.\ref{fig:intro}.

\begin{figure*}[t]
\centering
\includegraphics[width=1.0\linewidth]{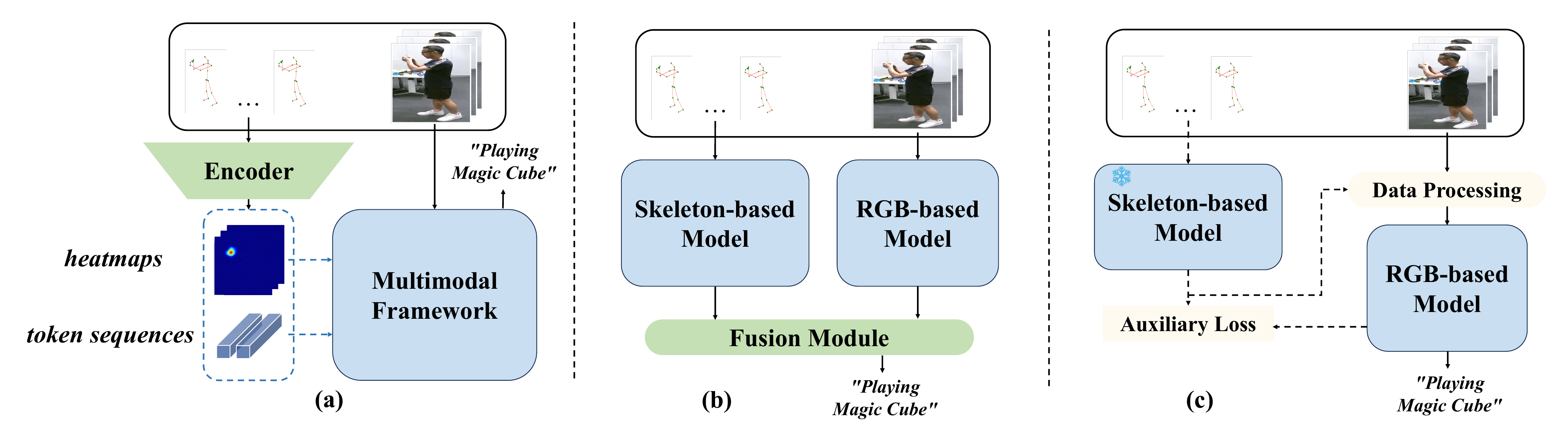}
\caption{\textbf{Primary directions of integrating RGB and skeleton modalities.}}
\label{fig:direction}
\end{figure*}

The core challenge of integrating the two modalities is their inherent heterogeneity: skeletal data is typically represented as an irregular graph, while RGB frames are represented on regular grids. As illustrated in Figure \ref{fig:direction}, previous studies have explored three primary directions: (1) encoding skeletal data into heatmaps or token sequences, thus transforming the task into learning representations from two homogeneous modalities \cite{kim2023crossmodal, duan2022revisiting, ahn2023star}; (2) extracting representations of the two modalities in separate pathways with late fusion \cite{das2020vpn, das2021vpn++}; (3) leveraging skeletal information as auxiliary prior knowledge to guide the data processing or model training of RGB-based methods \cite{bruce2022mmnet, reilly2024just}. 

Despite considerable success, they either neglect the inherent structure of skeletal data or fail to effectively exploit the correlation between skeletal and RGB data. Moreover, although the second direction shows promise in exploring feature correlations between the two modalities, the high computational cost of video models often results in a mismatch in temporal resolution. Typically, the RGB pathway operates on 8 frames, while the skeletal pathway can process more than 32 frames. Such temporal mismatch hinders the effective fusion of RGB and skeleton modalities.

Recent studies \cite{reilly2024just} have found that distinguishable visual cues are often located around human skeleton joints. Meanwhile, numerous works \cite{han2022vision, munir2023mobilevig, munir2024greedyvig} in the field of vision graph neural networks (ViGs) have demonstrated the effectiveness of encoding RGB image patches into graph representations. This observation naturally brings us to wonder: \textit{Could RGB frames adopt the same graph modeling paradigm of skeletal data ?} 

Motivated by this, we propose \textbf{PAN}, the first human-centric graph representation learning framework for multimodal action recognition. PAN utilizes visual foundation models to encode RGB frames extracted from videos. The corresponding 2D skeletal data is gathered synchronously to ensure consistent temporal alignment between the two modalities. Subsequently, token embeddings of patches containing human joints are sampled based on the joint coordinates provided by the 2D skeletal data. To reduce the dependency on high-quality skeletal data, we further propose attention-based post calibration to refine the sampled token embeddings. After sampling and post calibration, visual token embeddings are represented as spatiotemporal graphs (see Figure \ref{fig:intro}) and fed into GCNs.

Compared to previous works, PAN learns human-centric graph representations for RGB frames, providing a new perspective for integrating RGB and skeleton modalities. As temporal dependencies are modeled only among the sampled token embeddings within GCNs, PAN can utilize advanced visual foundation models while maintaining temporal alignment with skeleton-based methods. Furthermore, since both the visual token graphs in PAN and the skeletal graphs in skeleton-based methods are constructed based on the topology of human joints, they are structurally consistent and semantically aligned, bringing the potential for fine-grained cross-modal fusion. In this work, we present two preliminary variants: \textbf{PAN-Ensemble}, which employs dual-path GCNs followed by late fusion, and \textbf{PAN-Unified}, which performs unified graph representation learning within a single GCN. 

Our contributions can be summarized as follows:

\vspace{0.5em}
\begin{itemize}
\item To the best of our knowledge, PAN is the first human-centric graph representation learning framework for multimodal action recognition. By modeling RGB frames as human-centric spatiotemporal graphs, PAN offers a novel perspective for integrating RGB and skeleton modalities.

\item As PAN involves sampling token embeddings of patches containing human joints, we propose attention-based post calibration to reduce the dependency on high-quality 2D skeletal data by refining the sampled token embeddings.

\item Extensive experiments are conducted on three multimodal action recognition datasets. Both PAN-Ensemble and PAN-Unified achieve state-of-the-art (SOTA) performance in their respective settings of multimodal fusion: separate and unified modeling, respectively.
\end{itemize}

\section{Related Work}
\label{sec:relevant_work}

\subsection{Unimodal Action Recognition}

RGB-based methods and skeleton-based methods are two distinct groups for unimodal action recognition. Traditionally, 3DCNNs \cite{carreira2017quo, feichtenhofer2020x3d, lin2019tsm} and two-stream CNNs \cite{feichtenhofer2016convolutional, feichtenhofer2019slowfast, simonyan2014two} are utilized to process RGB video data. In recent years, with the advancement of vision transformers \cite{dosovitskiy2020image, fan2021multiscale, liu2021swin}, a variety of video transformer-based architectures \cite{arnab2021vivit, bertasius2021space, liu2022video} have been proposed for action recognition, demonstrating strong capabilities in capturing long-range spatiotemporal dependencies. For skeleton-based action recognition, since the induction of ST-GCN \cite{yan2018spatial}, GCNs \cite{li2019actional, shi2019two, cheng2020skeleton, chen2021channel, song2022constructing, liu2020disentangling} have emerged as a widely adopted solution due to their suitability for structured skeletal data. Recently, a few works \cite{wang20233mformer, zhou2022hypergraph, qu2024llms} have explored the adoption of transformer-based architectures or LLMs in this domain, suggesting potential directions for future research.

\subsection{Multimodal Action Recognition}

Given the limitations of individual modalities in capturing comprehensive spatiotemporal features, a growing number of methods have been proposed to enhance the robustness and generalization of action recognition models by combining RGB and skeleton modalities. PoseC3D \cite{duan2022revisiting} encodes 2D skeletal data into heatmaps, which are then fed into a 3D CNN. STAR-Transformer \cite{ahn2023star} and 3D deformable transformer \cite{kim2023crossmodal} represent skeletal data as token sequences and introduce specialized spatiotemporal attention mechanisms. VPN \cite{das2020vpn} employs stacked GCNs and a 3DCNN-based RGB backbone to extract embeddings from 3D skeletal data and RGB videos, respectively, which are then fused through a spatiotemporal coupler. VPN++ \cite{das2021vpn++} extends VPN by introducing feature and attention level distillations. More recently, $\pi$-ViT \cite{reilly2024just} augments the RGB representations through auxiliary loss terms induced by 2D and 3D skeletal data.

Among these methods, MMNet \cite{bruce2022mmnet} is the closest to our work. It builds a spatiotemporal region of interest feature map for each RGB video based on the embeddings of skeletal data, which is then fed into a 2DCNN. While both MMNet and our work avoid the use of video-based models, they differ in their underlying motivations. The intention of MMNet for this is to avoid the huge computational cost of video-based models, while ours is to address the temporal mismatch between the two modalities. In addition, our work samples visual token embeddings rather than raw images and conducts graph modeling, which represents a significant departure from MMNet.

\subsection{Vision Graph Neural Networks}

ViG-based architectures have been proven to be efficient computer vision backbones since the introduction of ViG \cite{han2022vision}, which constructs the graph through representing each token as a node. Vision Hyper-Graph Neural Network \cite{han2023vision} introduces the hypergraph structure to overcome the limitations of the original ViG. Works such as MobileViG \cite{munir2023mobilevig} and GreedyViG \cite{munir2024greedyvig} improve upon the graph construction strategy to reduce computational overhead while capturing robust vision representations. In contrast, our work leverages cross-modal guidance to sample crucial visual token embeddings from RGB frames before graph modeling and initializes the graph with prior knowledge (the physical connections of human joints), rather than constructing graphs from all available tokens.

\section{Methodology}

In this section, we present our human-centric graph representation learning framework for multimodal action recognition, PAN, and its variants that integrate PAN with skeleton-based methods. More precisely, Section 3.1 describes how PAN encodes raw RGB frames into spatiotemporal graphs, specifically the sampling of token embeddings. Section 3.2 introduces the proposed attention-based post calibration. Section 3.3 describes the employed basic blocks of GCNs and outlines the total architecture of PAN. Finally, Section 3.4 explains the two variants, which explore cross-modal fusion of RGB and skeleton modalities under a consistent human-centric graph modeling paradigm.

\begin{algorithm}[!b]
\SetAlFnt{\small}

\textbf{Notation:}

$M$: the number of persons; $J$: the number of joints; $T$: the number of frames; $H,W$: image size; $P$: patch size

\textbf{Input:} 

$\mathbf{S}  \in \mathbb{R}^{T \times M \times J \times 2}$ : 2D skeletal data

$\mathbf{R}  \in \mathbb{R}^{T \times \lceil\frac{H}{P}\rceil \times \lceil\frac{W}{P}\rceil \times C}$: visual token embeddings

\textbf{Procedure:} 

\hspace{0.5em}
\For{$t = 1$ to $T$}{
    \For{$m = 1$ to $M$}{
        \For{$j = 1$ to $J$}{
            Derive the index $\mathbf{I}_{t,m,j}$ as:
            $$ \mathbf{I}_{t,m,j} = \left\lfloor \frac{\mathbf{S}_{t,m,j}[1]}{P} \right\rfloor * \frac{W}{P} + \left\lfloor \frac{\mathbf{S}_{t,m,j}[0]}{P} \right\rfloor $$
            Sample patch from $\mathbf{R}_t$ based on $\mathbf{I}_{t,m,j}$
        }
    }
    Append sampled patch to $\mathbf{O}_t$
}

\hspace{0.5em}
Stack all $\mathbf{O}_t$ into a tensor $\mathbf{O} \in\mathbb{R}^{T\times M \times J \times C}$

\textbf{Output:} 
$\mathbf{O}$: sampled token embeddings
\caption{Guided Token Sampling}
\label{sampling}
\end{algorithm}

\subsection{Visual Token Graph Encoder}

\begin{figure*}[t]
\centering
\includegraphics[width=1.0\linewidth]{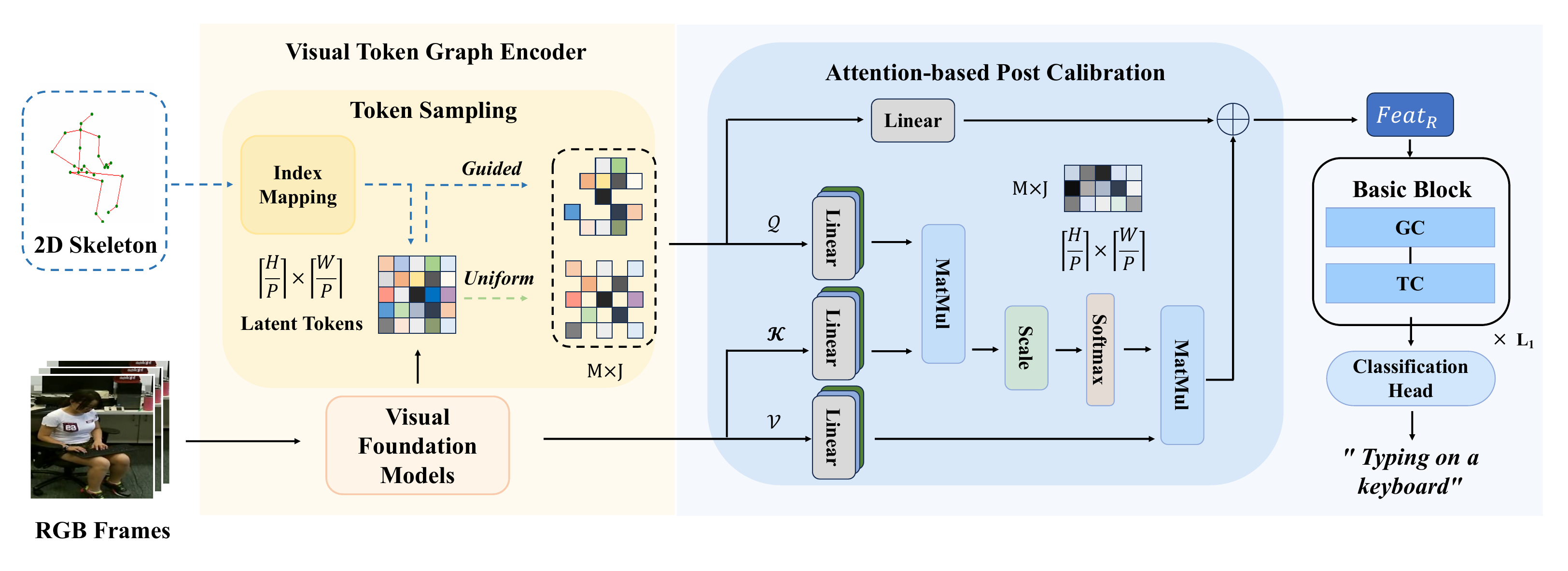}
\caption{\textbf{Overview of PAN.} 2D skeletal data and the index mapping are only required for guided sampling, whereas uniform sampling relies solely on RGB frames. In the attention-based post calibration, the sampled and original token embeddings serve as \textit{query} and \textit{key / value} respectively. $L_1$ basic blocks of GCN are then stacked, followed by a classification head. Here GC and TC denote graph convolution and temporal convolution, respectively.}
\label{fig:rgbStream}
\end{figure*}

Given an input video $V$, we first extract RGB frames with the shape $T\times H\times W\times 3$. As we model temporal dependencies between visual token graphs rather than directly between raw RGB frames, the number of extracted frames $T$ can be quite large to align with skeleton-based methods. The extracted frames are then fed into ViT-based visual foundation models, where each frame is decomposed into $\lceil \frac{H}{P}\rceil \times \lceil \frac{W}{P}\rceil$ patches and subsequently tokenized. A $cls$ token is concatenated to the token sequence, resulting in a total of $\lceil \frac{H}{P}\rceil \times \lceil \frac{W}{P}\rceil + 1$ tokens. Typically, after being processed by a series of transformer blocks, the output $cls$ token is pooled out for classification. In contrast, we discard it to preserve the spatial resolution and the remaining $\lceil \frac{H}{P}\rceil \times \lceil \frac{W}{P}\rceil$ tokens are used for sampling.

In PAN, we investigate two distinct sampling strategies. For guided sampling, visual token embeddings are sampled based on 2D skeletal data, as illustrated in Algorithm \ref{sampling}. A simple index mapping is applied to establish the correspondence between the 2D skeleton joints and the visual tokens. The sampled token embeddings, denoted by $\mathbf{O}\in\mathbb{R}^{T\times M \times J \times C}$, represent visual semantic features around human joints and are structured into visual token graphs, where nodes correspond to joint-specific visual tokens and edges encode spatial or temporal relationships among them. For uniform sampling, we employ linear interpolation to select $J$ tokens from visual token embeddings and simply replicate them for each person.

\subsection{Attention-based Post Calibration}

Since the index mapping in guided sampling relies on the joint coordinates provided by the 2D skeletal data, it could be vulnerable to inaccuracies in 2D pose estimation. To mitigate the issue, we propose attention-based post calibration that refines sampled token embeddings in a low-dimensional latent space. Given the original token embeddings $\mathbf{R}$ and the sampled token embeddings $\mathbf{O}$, we linearly project $\mathbf{O}$ into queries and $\mathbf{R}$ into keys and values of $C_R$ dimensions with learned matrices $\textbf{W}_Q,\textbf{W}_K,\textbf{W}_V \in \mathbb{R}^{C \times C_R}$ and compute a cross-attention map as:

\begin{equation}
    \text{Attn}=\text{softmax}\left(\frac{\mathbf{R} \mathbf{W}_K (\mathbf{O} \mathbf{W}_Q)^T}{\sqrt{C_R}}\right).
\end{equation}

In practice, $H$ parallel heads are employed to learn from multiple representation subspaces. The refined visual token embeddings $\text{Feat}_{\text{R}}$ are formulated as:
\begin{equation}
    \text{Feat}_{\text{R}} =  \text{Concat}\left(\text{Attn}_1, ..., \text{Attn}_H\right) \mathbf{R}\mathbf{W}_V + \mathbf{O}\mathbf{W}_{res},
\end{equation}
where $\text{Feat}_{\text{R}}\in \mathbb{R}^{T\times M\times J\times C_R}$ is the refined token embeddings and $\mathbf{W}_{res}\in \mathbb{R}^{C\times C_R}$ is the residual projection matrix.

\subsection{PAN Architecture}

After the visual token graph encoder and attention-based post calibration, multiple basic blocks of GCN are stacked to process $\text{Feat}_{\text{R}}$. Here, we first revisit the formulations for vanilla graph convolution for skeleton-based action recognition. Typically, the human body within a motion sequence is represented as a spatio-temporal graph. The graph is denoted as $\mathcal{G} = (\mathcal{V}, \mathcal{E})$, where $\mathcal{V} = \{ v_1, v_2, \dots, v_J \}$ is the set of $J$ vertices representing joints and $\mathcal{E}$ is the edge set representing the correlations between joints. Typically, $\mathcal{G}$ is formulated by $ \mathbf{H}\in\mathbb{R}^{J \times T \times C}$ and $\mathbf{A}\in\mathbb{R}^{J \times J} $. $\mathbf{H}$ is the feature tensor of $J$ vertices across $T$ frames, and $v_i$'s feature at frame $t$ is denoted as $x_{i,t}\in\mathbb{R}^{C}$. $\mathbf{A}$ is the adjacency matrix, with its elements $a_{ij}$ representing the correlation between $v_{i}$ and $v_{j}$. The vanilla graph convolution proposed by \cite{kipf2016semi} is widely adopted:
\begin{equation}
	  \mathbf{H}^{(l+1)}=\sigma(\hat{\mathbf{A}}\mathbf{H}^{(l)}\mathbf{W}^{(l)}),
	  \label{eq:vanillagc}
\end{equation}
where $\mathbf{H}^{(l)}\in\mathbb{R}^{J \times T \times d^{(l)}}$ denotes hidden feature representation of the $l$-th layer, $\mathbf{W}^{(l)}\in\mathbb{R}^{d^{(l)} \times d^{(l + 1)}}$ is the learnable parameter matrix of the $l$-th layer, $\hat{\mathbf{A}}\in\mathbb{R}^{J \times J}$ is the normalized adjacency matrix and $\sigma$ indicates activation function. For multi-person samples, each individual's features are modeled independently, and their predicted class scores are aggregated to produce the final prediction.

In PAN, $\text{Feat}_{\text{R}}$ represents human-centric visual semantic features around joints and aligns with the structure of $\mathcal{G} = (\mathcal{V}, \mathcal{E})$. This alignment allows it to seamlessly serve as input to existing GCN-based methods. Hence, we adopt the basic block in CTR-GCN \cite{chen2021channel} due to its stability. 

The block consists of channel-wise topology refinement graph convolution (CTR-GC) and multi-scale temporal convolution (MS-TC). Given the input feature representations $\mathbf{H}^{(l)}\in\mathbb{R}^{J \times T \times d^{(l)}}$ of the $l$-th layer, CTR-GC utilizes a correlation modeling function $\mathcal{M}$ to calculate the correlation weights between nodes, which can be formulated as:
\begin{equation}
	  \mathcal{M}(\mathbf{H}^{(l)}) = \xi(\psi(\mathbf{H}^{(l)}) - \phi(\mathbf{H}^{(l)})),
	\label{eq:transform}
\end{equation}
where $\psi(\mathbf{H}^{(l)})\in\mathbb{R}^{d' \times J \times 1}$ and $\phi(\mathbf{H}^{(l)})\in\mathbb{R}^{d' \times 1 \times J}$ are two distinct feature embeddings implemented by 1x1 convolution. $\mathcal{M}(\mathbf{H}^{(l)})\in\mathbb{R}^{d' \times J \times J}$ represents the data-dependent correlation weights. 

The output feature of CTR-GC can be formulated as:
\begin{equation}
\label{topo_add}
\mathbf{H}^{(l+1)} =  \left(\lambda * \mathcal{M}(\mathbf{H}^{(l)}) + \mathbf{\tilde{A}}\right)\mathbf{H}^{(l)}\mathbf{W}^{(l)},
\end{equation}
where $\mathbf{\tilde{A}} \in\mathbb{R}^{J \times J}$ is a learnable adjacency matrix initialized with the physical connections of human joints. $\lambda$ is a learnable weighting factor to balance the data-dependent channel-wise correlation weights and the shared weights across channels.  

MS-TC contains four branches, each starting with a $1\times 1$ convolution for channel projection. The first two branches utilize a 1D temporal convolution with different dilation rates after $1\times 1$ convolution, while the third branch employs a $\text{MaxPooling}$ operation. The output of four branches is concatenated. In general, the block can be formulated as:
\begin{equation}
    \mathbf{H}^{(l+1)} = \mathbf{H}^{(l)} + \sigma\left(\sum^G\text{CTR-GC}(\mathbf{H}^{(l)})\right)
\end{equation}
\begin{equation}
    \mathbf{H}^{(l+1)} = \text{MS-TC}(\mathbf{H}^{(l+1)}) + \mathbf{H}^{(l)},
\end{equation}
where $G$ is the number of groups in CTR-GC. PAN follows \cite{chen2021channel} in hyperparameter settings as $G=3$ and $d'=\frac{1}{8}d^{(l)}$.

\begin{figure}[t]
\centering
\includegraphics[width=1.0\linewidth]{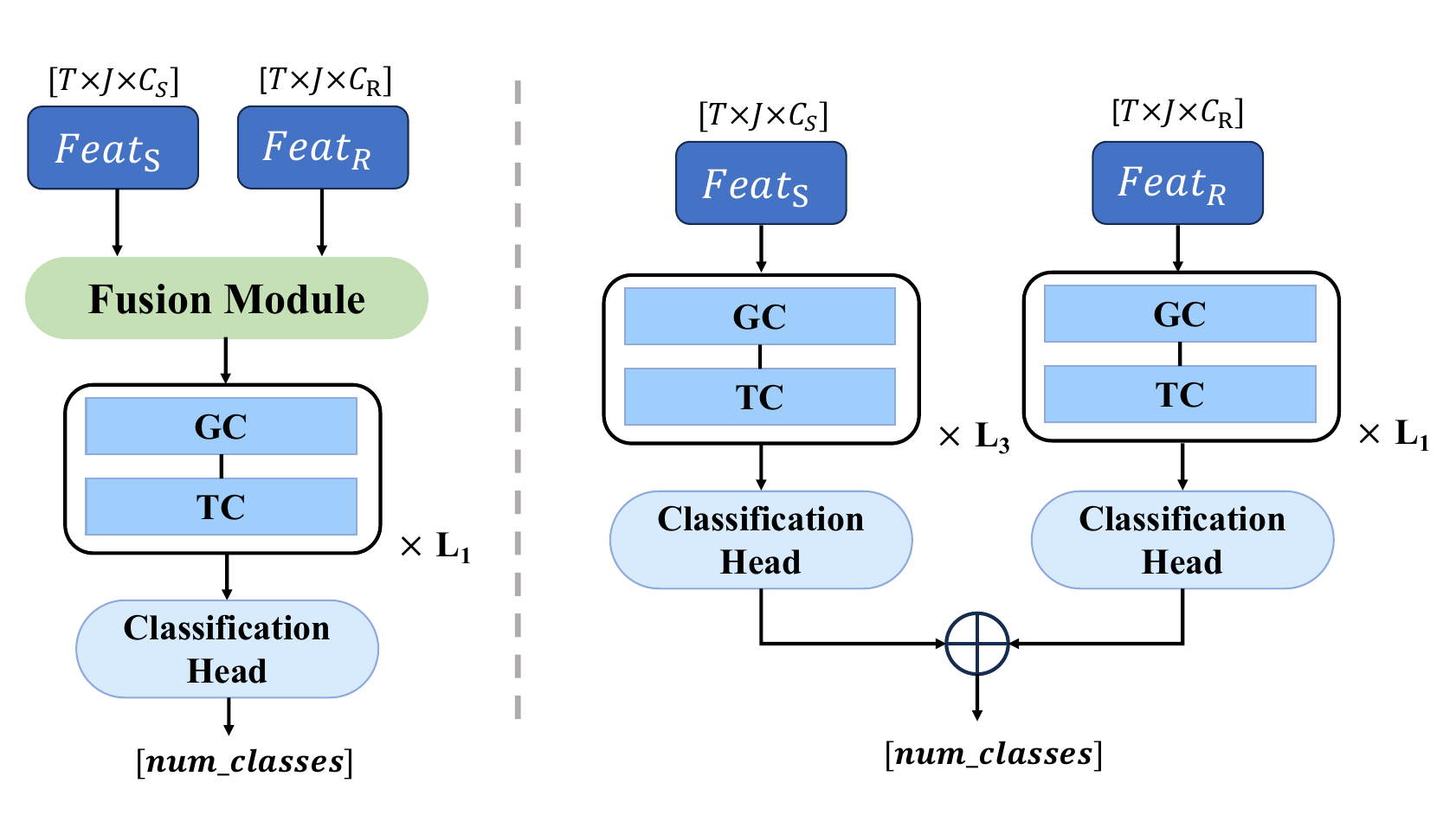}
\caption{\textbf{Left:} PAN-Unified, where visual token graph embeddings and skeletal graph embeddings are modeled in a single GCN. \textbf{Right:} PAN-Ensemble, in which late fusion is applied by summing the classification scores.}
\label{fig:variants}
\end{figure}

An overview of PAN is illustrated in Figure \ref{fig:rgbStream}. For guided sampling, the input RGB frames and 2D skeletal data are gathered synchronously. The number of sampled frames is set to $T=32$. In the attention-based post calibration, we adopt the setting of $H=4$ and the output channel $C_R=256$. After the visual token graph encoder and the attention-based post calibration, $L_1=5$ basic blocks are stacked and their input channels are 256-256-256-512-512. The classification head is then applied to generate predictions, which consists of a global average pooling operation, a fully connected layer, and a softmax classifier.

\subsection{Variants of PAN for Cross-Modal Fusion}

Since visual token graph embeddings $\text{Feat}_{\text{R}}$ and skeletal graph embeddings are structurally consistent and semantically aligned, their fusion is worth exploring. To this end, we present two preliminary variants of PAN: PAN-Ensemble and PAN-Unified, as illustrated in Figure \ref{fig:variants}.

First, as $\text{Feat}_{\text{R}}$ originates from a high-dimensional latent space and encodes higher-level semantics than raw 3D skeletal data, we extract skeletal graph embeddings $\text{Feat}_{\text{S}}\in\mathbb{R}^{T \times J \times C_S}$ through $L_2$ basic blocks. In PAN-Ensemble, $\text{Feat}_{\text{S}}$ and $\text{Feat}_{\text{R}}$ are then processed in separate pathways, followed by a late fusion. Unlike previous approaches that typically employ heterogeneous modeling strategies for different pathways, the two pathways in PAN-Ensemble adopt the same modeling paradigm, consisting of $L_1$ and $L_3$ basic blocks. The final prediction is obtained by summing the classification scores from both pathways. In PAN-Unified, $\text{Feat}_{\text{S}}$ and $\text{Feat}_{\text{R}}$ are fused before being processed through a single GCN, which consists of $L_1$ basic blocks. The fusion module is implemented through a $1\times 1$ convolution that aligns feature dimension from $C_S$ to $C_R$, followed by an element-wise summation between the visual token embeddings and the processed skeletal graph embeddings.
\section{Experiments}

\subsection{Datasets}
\noindent \textbf{NTU RGB+D} \cite{shahroudy2016ntu} is a large-scale, multi-modal benchmark for human action recognition, comprising 56,880 action samples across 60 distinct classes. The action samples are performed by 40 participants in different age groups and recorded from 3 camera views. For each sample, synchronized RGB videos, depth maps, and 2D/3D skeleton sequences extracted using Microsoft Kinect v2 sensors are provided. We report the top-1 classification accuracy under cross-subject (CS) and cross-view (CV) protocols.

\noindent \textbf{NTU RGB+D 120} \cite{liu2019ntu} is an extended version of NTU RGB+D with a total of 114480 samples across 120 action classes. It was performed by 106 volunteers under 32 distinct setups for locations and backgrounds. The dataset introduces two challenging evaluation protocols: cross-subject (Csub) and cross-setup (CSet).

\noindent \textbf{Tokoyo-Smarthome} \cite{das2019toyota} consists of 16115 action samples across 31 action classes, which are performed by 18 elderly individuals under 7 camera views. The 2D/3D skeleton sequences are extracted through LCRNet \cite{rogez2019lcr}. Due to the unbalanced nature of the dataset, we report the mean class-accuracy (mCA) under cross-subject (CS) and cross-view (CV1, CV2) settings for evaluation.

\begin{table}[t]
\centering
\caption{Comparison with SOTA on NTU120. Top-1 classification accuracies are reported. $\circ$ indicates that the modality has been used only in training. $\dagger$ indicates that re-estimated data has been used instead of original annotations. Bold text indicates best performance, underline indicates second best performance. All results are taken from official sources or \cite{reilly2024just}.}
\label{tab:ntu120}
\resizebox{0.48\textwidth}{!}{
\begin{tabular}{lcccrr}
\toprule
\multirow{2}{*}{\textbf{Methods}} & \multicolumn{3}{c}{\textbf{Modality}} & \multirow{2}{*}{\textbf{CSub}} & \multirow{2}{*}{\textbf{CSet}} \\
& \textbf{2D Ske} & \textbf{3D Ske} & \textbf{RGB} & & \\
\midrule

\rowcolor{gray!30}
\multicolumn{6}{l}{\textit{Unimodal Methods}} \\
CTR-GCN\cite{chen2021channel}& \ding{55}& \ding{51}& \ding{55}& 88.9& 90.4 \\
InfoGCN\cite{chi2022infogcn}& \ding{55}& \ding{51}& \ding{55}& 89.8& 91.2 \\
BlockGCN\cite{zhou2024blockgcn}& \ding{55}& \ding{51}& \ding{55}& 90.3& 91.5 \\
ProtoGCN\cite{liu2025revealing}& \ding{55}& \ding{51}& \ding{55}& 90.9& \underline{92.2} \\
3Mformer\cite{wang20233mformer}& \ding{55}& \ding{51}& \ding{55}& \underline{92.0}& \textbf{93.8} \\

MotionFormer\cite{patrick2021keeping}& \ding{55}& \ding{55}& \ding{51}& 87.0& 87.9 \\
TimeSformer\cite{bertasius2021space}& \ding{55}& \ding{55}& \ding{51}& 90.6& 91.6 \\
Video Swin\cite{liu2022video}& \ding{55}& \ding{55}& \ding{51}& 91.4& 92.1 \\

\rowcolor{green!20}
PAN (Even)& \ding{55}& \ding{55}& \ding{51}& \textbf{93.5}& \textbf{93.8} \\

\midrule
\rowcolor{gray!30}
\multicolumn{6}{l}{\textit{Multimodal Methods(separate modeling)}} \\

VPN++ \cite{das2021vpn++}& \ding{55} & $\circ$ & \ding{51}& 86.7& 89.3 \\
VPN++ \cite{das2021vpn++} (late fusion)& \ding{55} & \ding{51} & \ding{51}& 90.7& 92.5 \\
$\pi$-ViT \cite{reilly2024just}& $\circ$ & $\circ$ & \ding{51}& 91.9& 92.9 \\
$\pi$-ViT \cite{reilly2024just} (late fusion) & $\circ$ & \ding{51} & \ding{51}& 95.1& 96.1 \\
PoseC3D \cite{duan2022revisiting}& $\dagger$ & \ding{55} & \ding{51}& \underline{95.3}& \underline{96.4} \\

\rowcolor{green!20}
PAN (Guided)& \ding{51}& \ding{55}& \ding{51}& 93.6& 93.9 \\
\rowcolor{green!20}
PAN-Ensemble& \ding{51}& \ding{51}& \ding{51}& \textbf{96.2}& \textbf{97.0} \\

\midrule
\rowcolor{gray!30}
\multicolumn{6}{l}{\textit{Multimodal Methods(unified modeling)}} \\
STAR-Transformer \cite{ahn2023star}& \ding{51} & \ding{55} & \ding{51}& 90.3& \underline{92.7} \\
3D-Def-Transformer \cite{kim2023crossmodal}& \ding{51} & \ding{55} & \ding{51}& \underline{90.5}& 91.4 \\

\rowcolor{green!20}
PAN-Unified& \ding{51}& \ding{51}& \ding{51}& \textbf{94.1}& \textbf{93.0} \\

\bottomrule
\end{tabular}}
\end{table}
\subsection{Implementation Details}
\label{section:imple-detail}
In all experiments except for the ablation studies presented in Table \ref{VFM}, DINOv2-small \cite{oquab2023dinov2} is employed to encode RGB frames within the visual token graph encoder. We set $L_1 = L_2 = L_3 = 5$ in accordance with the original structure design of CTR-GCN \cite{chen2021channel}. The RGB inputs to our model are video frames of size $32\times 224\times 224$.  Frames are sampled at a rate of $\frac{1}{32}$ for Smarthome and uniform sampling is used for NTU60 and NTU120. If the number of frames in a sample is less than 32, duplicated frames are used to meet the required length. Notably, we follow $\pi$-ViT \cite{reilly2024just} and use tracks of human crops as our input RGB frames. More details are available in the supplementary material. For 3D skeleton sequences, the temporal length is initially 64 and halved to 32 after $L_2$ basic blocks to align with the visual token embeddings.

All experiments are conducted on 4 Tesla V100 GPUs. The cross-entropy loss is used for optimization, and the SGD optimizer is adopted with a momentum of 0.9 and a weight decay of 4e-4. The learning rate is initialized at 0.1 and decayed by a factor of 10 at epochs 35 and 55, with a total training duration of 65 epochs. 

\begin{table}[t]
\centering
\caption{Comparison with SOTA on NTU60. Top-1 classification accuracies are reported. $\circ$ indicates that the modality has been used only in training. $\dagger$ indicates that re-estimated data has been used instead of original annotations. Bold text indicates best performance, underline indicates second best performance. All results are taken from official sources or \cite{reilly2024just}.}
\label{tab:ntu60}
\resizebox{0.48\textwidth}{!}{
\begin{tabular}{lcccrr}
\toprule
\multirow{2}{*}{\textbf{Methods}} & \multicolumn{3}{c}{\textbf{Modality}} & \multirow{2}{*}{\textbf{CS}} & \multirow{2}{*}{\textbf{CV}} \\
& \textbf{2D Ske} & \textbf{3D Ske} & \textbf{RGB} & & \\
\midrule

\rowcolor{gray!30}
\multicolumn{6}{l}{\textit{Unimodal Methods}} \\
CTR-GCN\cite{chen2021channel}& \ding{55}& \ding{51}& \ding{55}& 92.4& 96.4 \\
InfoGCN\cite{chi2022infogcn}& \ding{55}& \ding{51}& \ding{55}& 93.0& 97.1 \\
BlockGCN\cite{zhou2024blockgcn}& \ding{55}& \ding{51}& \ding{55}& 93.1& 97.0 \\
ProtoGCN\cite{liu2025revealing}& \ding{55}& \ding{51}& \ding{55}& 93.8& 97.8 \\
3Mformer\cite{wang20233mformer}& \ding{55}& \ding{51}& \ding{55}& \underline{94.8}& \textbf{98.7} \\

MotionFormer\cite{patrick2021keeping}& \ding{55}& \ding{55}& \ding{51}& 85.7& 91.6 \\
TimeSformer\cite{bertasius2021space}& \ding{55}& \ding{55}& \ding{51}& 93.0& 97.2 \\
Video Swin\cite{liu2022video}& \ding{55}& \ding{55}& \ding{51}& 93.4& 96.6 \\

\rowcolor{green!20}
PAN (Even)& \ding{55}& \ding{55}& \ding{51}& \textbf{95.5}& \underline{98.5} \\

\midrule
\rowcolor{gray!30}
\multicolumn{6}{l}{\textit{Multimodal Methods(separate modeling)}} \\

VPN++ \cite{das2021vpn++}& \ding{55} & $\circ$ & \ding{51}& 93.5& 96.1\\
VPN++ \cite{das2021vpn++} (late fusion)& \ding{55} & \ding{51} & \ding{51}& 96.6& 99.1 \\
$\pi$-ViT \cite{reilly2024just}& $\circ$ & $\circ$ & \ding{51}& 94.0& 97.9 \\
$\pi$-ViT \cite{reilly2024just} (late fusion) & $\circ$ & \ding{51} & \ding{51}& 96.3& 99.0 \\
PoseC3D \cite{duan2022revisiting}& $\dagger$ & \ding{55} & \ding{51}& \underline{97.0}& \textbf{99.6} \\

\rowcolor{green!20}
PAN (Guided)& \ding{51}& \ding{55}& \ding{51}& 95.7& 98.6 \\
\rowcolor{green!20}
PAN-Ensemble& \ding{51}& \ding{51}& \ding{51}& \textbf{97.4}& \underline{99.5} \\

\midrule
\rowcolor{gray!30}
\multicolumn{6}{l}{\textit{Multimodal Methods(unified modeling)}} \\
STAR-Transformer \cite{ahn2023star}& \ding{51} & \ding{55} & \ding{51}& 92.0& 96.5 \\
3D-Def-Transformer \cite{kim2023crossmodal}& \ding{51} & \ding{55} & \ding{51}& \underline{94.3}& 97.9 \\

\rowcolor{green!20}
PAN-Unified& \ding{51}& \ding{51}& \ding{51}& \textbf{96.3}& \textbf{98.9} \\

\bottomrule
\end{tabular}}
\end{table}

\subsection{Comparison with State-of-the-art}

Since PAN can be categorized as either an RGB-based method or a multimodal method depending on the adopted sampling strategy, we compare (1) PAN (Even) with unimodal methods and (2) PAN (Guided) with multimodal methods that adopt separate modeling. In addition, PAN-Ensemble and PAN-Unified are evaluated against multimodal methods adopting separate modeling and unified modeling, respectively.

\noindent \textbf{NTU60 and NTU120.} We present the comparison between our method and the SOTA on NTU60 and NTU120 in Table \ref{tab:ntu60} and Table \ref{tab:ntu120}, respectively. Since the NTU datasets involve up to two persons, we pad zero for samples containing only one person following \cite{chen2021channel}. Due to the proposed attention-based post calibration, PAN (Even) demonstrates high robustness, with a maximum performance drop of only 0.2\% compared to PAN (Guided). On NTU120, PAN-Ensemble exhibits an improvement of 0.9\% and 0.6\% respectively over the previous SOTA. In contrast, on NTU60, the performance gain is marginal. Our PAN-Unified outperforms the previous SOTA methods adopting unified modeling across all benchmarks.

\noindent \textbf{Smarthome.} In Table \ref{tab:smarthome}, we compare our method with the SOTA on Smarthome. As tokens containing human joints account for only 5.86\% of the original visual tokens, we double the number of sampled tokens (set $M=2$) through zero padding or token replication. The two strategies are evaluated within PAN (Even) and token replication is adopted as the final strategy due to its overall performance across three benchmarks. While PAN (Even) achieves SOTA among unimodal approaches,  PAN (Guided) exhibits a performance degradation of -0.9\%, -1.4\% and -1.8\% on the CS, CV1 and CV2 settings, respectively. This is primarily due to the low quality of skeletal data in real-world scenarios. Consequently, for Smarthome dataset, we build PAN-Ensemble and PAN-Unified based on PAN(Even). Among multimodal methods that adopt separate modeling, PAN-Ensemble achieves SOTA performance on the CS and CV2 protocols. Notably, a significant improvement of +8.3\% is achieved over the previous SOTA method $\pi$-ViT on the CV2 protocal.

\begin{table}[t]
\centering
\caption{Comparison with SOTA on Smarthome. Mean class accuracies are reported. $\circ$ indicates that the modality has been used only in training. $\dagger$ indicates that re-estimated data has been used instead of original annotations. $\star$ ans $\ast$ indicate that we pad zero or replicate the sampled tokens to extend $M$ to 2 within the visual token graph encoder, respectively. Bold text indicates best performance, underline indicates second best performance. All results are taken from official sources or \cite{reilly2024just}.}
\label{tab:smarthome}
\resizebox{0.48\textwidth}{!}{
\begin{tabular}{lcccrrr}
\toprule
\multirow{2}{*}{\textbf{Methods}} & \multicolumn{3}{c}{\textbf{Modality}} & \multirow{2}{*}{\textbf{CS}} & \multirow{2}{*}{\textbf{CV1}} & \multirow{2}{*}{\textbf{CV2}} \\
& \textbf{2D Ske} & \textbf{3D Ske} & \textbf{RGB} & & & \\
\midrule

\rowcolor{gray!30}
\multicolumn{7}{l}{\textit{Unimodal Methods}} \\
2s-AGCN\cite{shi2019two}& \ding{55}& \ding{51}& \ding{55}& 60.9& 21.6 & 32.3 \\
PoseC3D\cite{chi2022infogcn}& $\dagger$ & \ding{55}& \ding{55}& 50.6& 20.0& 28.2 \\
Hyperformer\cite{wang20233mformer}& \ding{55}& \ding{51}& \ding{55}& 57.5& 31.6 & 35.2 \\

MotionFormer\cite{patrick2021keeping}& \ding{55}& \ding{55}& \ding{51}& 65.8& 45.2& 51.0 \\
TimeSformer\cite{bertasius2021space}& \ding{55}& \ding{55}& \ding{51}& 68.4& \underline{50.0}& \underline{60.6} \\
Video Swin\cite{liu2022video}& \ding{55}& \ding{55}& \ding{51}& 69.8& 36.6& 48.6 \\

\rowcolor{green!20}
PAN (Even)$\star$ & \ding{55}& \ding{55}& \ding{51}& \underline{70.5}& \textbf{50.2} & 58.7\\
% 这一行是pad 0可能还加了额外归一化，但是
\rowcolor{green!20}
PAN (Even)$\ast$ & \ding{55}& \ding{55}& \ding{51}& \textbf{70.8}& 47.9 & \textbf{72.9}\\ 
% 这一行是重复

\midrule
\rowcolor{gray!30}
\multicolumn{7}{l}{\textit{Multimodal Methods(separate modeling)}} \\

VPN++ \cite{das2021vpn++}& \ding{55} & $\circ$ & \ding{51}& 69.0& -& 54.9\\
VPN++ \cite{das2021vpn++} (late fusion)& \ding{55} & \ding{51} & \ding{51}& 71.0& -& 58.1 \\
$\pi$-ViT \cite{reilly2024just}& $\circ$ & $\circ$ & \ding{51}& 72.9& \underline{55.2}& 64.8 \\
$\pi$-ViT \cite{reilly2024just} (late fusion) & $\circ$ & \ding{51} & \ding{51}& \textbf{73.1} & \textbf{55.6}& 65.0 \\
PoseC3D \cite{duan2022revisiting}& $\dagger$ & \ding{55} & \ding{51}& 53.8& 21.5& 33.4 \\

\rowcolor{green!20}
PAN (Guided)& \ding{51}& \ding{55}& \ding{51}& 69.9& 46.5& 71.1 \\
\rowcolor{green!20}
PAN-Ensemble& \ding{55}& \ding{51}& \ding{51}& \underline{73.0}& 49.2& \textbf{73.3} \\

\midrule
\rowcolor{gray!30}
\multicolumn{7}{l}{\textit{Multimodal Methods(unified modeling)}} \\

\rowcolor{green!20}
PAN-Unified& \ding{55}& \ding{51}& \ding{51}& 71.0 & 48.2& 72.1\\
\bottomrule
\end{tabular}}
\end{table}

\subsection{Ablation Study}

\begin{table}[t]
\centering
\caption{Ablation on visual foundation models. We report the number of learnable parameters in PAN.}
\label{VFM}
\resizebox{0.48\textwidth}{!}{
\begin{tabular}{l|cc|cc}
\toprule
\textbf{Model Type} & \textbf{Hidden Dim} & \textbf{Patch Size} & \textbf{Params (M)} & \textbf{Acc (\%)} \\
\midrule
Dinov2\cite{oquab2023dinov2}   & 384 & 14 & 5.54& \textbf{93.64} \\
MAE\cite{he2022masked} & 768 & 16 & 5.93& 92.05 \\
CLIP\cite{radford2021learning} & 768 & 16 & 5.93& 92.26 \\
\bottomrule
\end{tabular}
}
\end{table}

\begin{table}[t]
\centering
\begin{minipage}{0.48\linewidth}
\centering
\caption{Ablation on post calibration, sampling strategy and number of channels.}
\label{tab:calibration}
\resizebox{\textwidth}{!}{
\begin{tabular}{cccc}
\toprule
\multirow{2}{*}{\textbf{Strategy}} & \multirow{2}{*}{$C_R$} & \multicolumn{2}{c}{\textbf{Post Calibration}} \\
 &  & \ding{55} & \ding{51} \\ \midrule
Even  & 128 & 85.31 & 92.87 \\
Even  & 256 & 89.19 & 93.51 \\ 
\midrule
Guided  & 128 & 91.35 & 93.03 \\
Guided  & 256 & 92.27 & 93.64 \\
Guided  & 384 & 92.44 & 93.63 \\
\bottomrule
\end{tabular}
}
\end{minipage}
\hfill
\begin{minipage}{0.49\linewidth}
\centering
\caption{Ablation study of components in PAN. We report the number of learnable parameters.}
\label{tab:components}
\resizebox{\textwidth}{!}{
\begin{tabular}{ccc}
\toprule
\textbf{Model Type} & \textbf{Params (M)} & \textbf{Acc (\%)} \\
\midrule
Full & 5.54 & 93.64 \\
w/o calibration & 5.17 & 92.27$^{\textcolor{green}{-1.37\%}}$ \\
w/o GC & 3.11 & 92.91$^{\textcolor{green}{-0.73\%}}$ \\
w/o TC & 4.96 & 77.12$^{\textcolor{green}{-16.52\%}}$ \\
w/o PAN & 0.05 & 56.36$^{\textcolor{green}{-37.28\%}}$ \\
\bottomrule
\end{tabular}
}
\end{minipage}

\end{table}

In this section, we conduct ablation studies on the design of PAN and its variants on the NTU120 CSub protocal.

\noindent \textbf{Choice of visual foundation models.} We train our PAN (Guided) using self-supervised models including DINOv2 \cite{oquab2023dinov2} and MAE \cite{he2022masked}, and the image-text contrastive learning model CLIP \cite{radford2021learning}. For all cases we set $C_R$ to 256. The results are summarized in Table \ref{VFM}. PAN achieves SOTA performance with all these vision foundation models. Among them, the model trained with DINOv2 achieves the best performance, attaining a top-1 accuracy of 93.64\%.

\noindent \textbf{Effect of post calibration on sampling strategies.} In Table \ref{tab:calibration}, we investigate how different sampling strategies perform with and without post calibration. When post calibration is not applied, the guided sampling strategy consistently outperforms the even strategy by a considerable margin across various $C_R$ configurations. However, when post calibration is applied, the performance gap narrows significantly. For example, when post calibration is applied for $C_R=256$, the performance gap between guided and even sampling strategy decreases from 3.08\% to 0.13\%. This demonstrates that post calibration effectively mitigates the impact of the adopted sampling strategy on model performance, thereby reducing the dependency on high-quality 2D skeletal data.

\noindent \textbf{Effectiveness of graph modeling paradigm.} Since PAN applies the graph modeling paradigm originally used for skeletal data to RGB frames, we conduct ablation studies to evaluate its effectiveness, as shown in Table \ref{tab:components}. Specifically, 'Full' denotes Dinov2 + PAN (Guided), 'w/o calibration' denotes that the attention-based post calibration is removed, 'w/o GC' denotes that the graph convolutions within basic blocks are replaced by $1\times1$ 2D convolutions, 'w/o TC' denotes that the multi-scale temporal convolutions within basic blocks are replaced by a combination of temporal pooling and $1\times1$ 2D convolutions, and 'w/o PAN' denotes that the original visual token embeddings are directly fed into a classification head without any graph modeling. 

The results demonstrate that when both temporal dependencies and graph modeling are omitted ('w/o PAN'), the performance drops significantly to 56.36\%. When the visual token graph is encoded but not updated ('w/o GC'), the performance only drops 0.73\%. These findings highlight the importance of temporal dependencies as well as the effectiveness of our human-centric graph modeling paradigm.

\begin{figure}[t]
\centering
\includegraphics[width=1.0\linewidth]{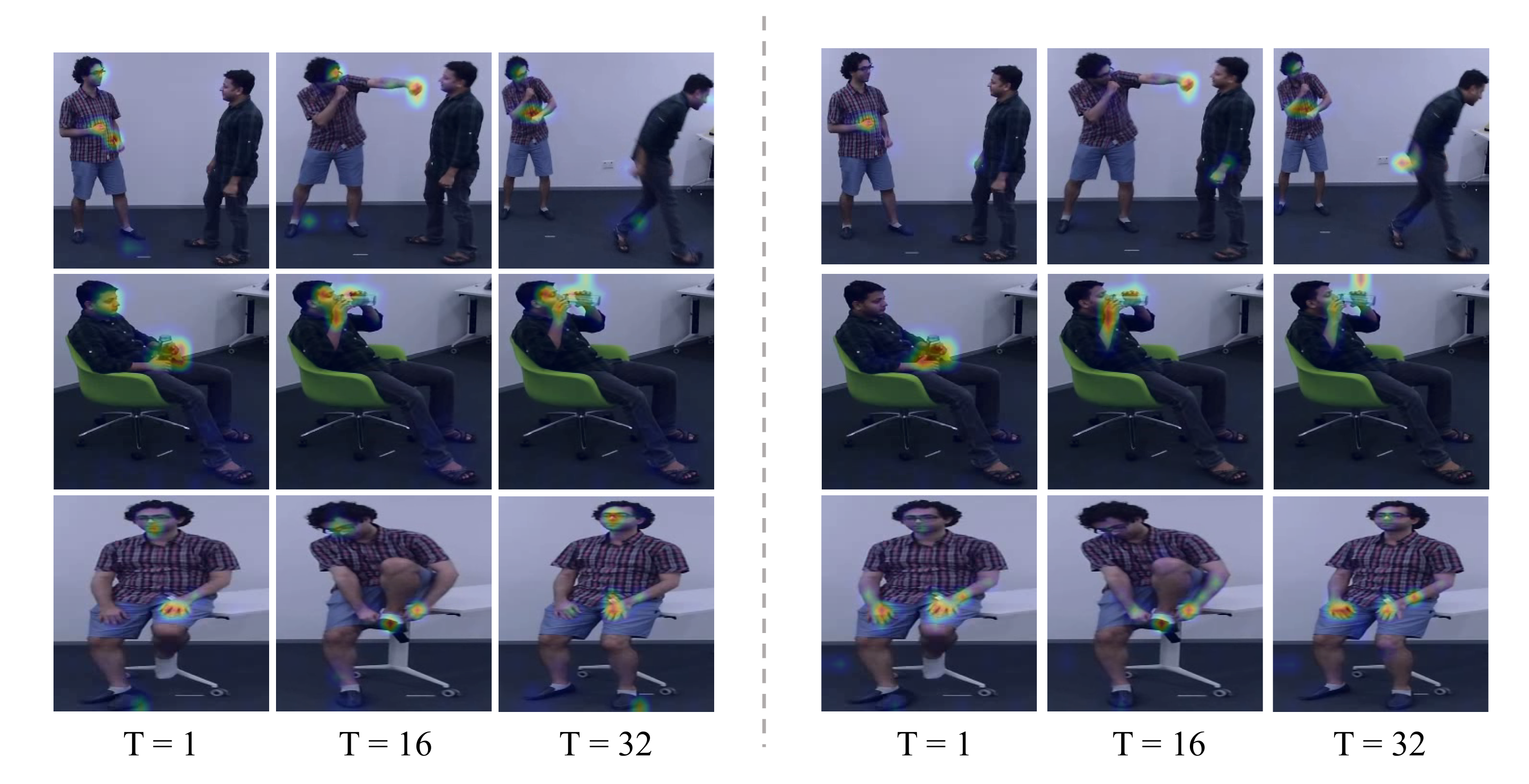}
\caption{\textbf{Visualization of attention maps} on three actions. \textbf{Left:} Guided Sampling; \textbf{Right:} Even Sampling. The actions are punching other person, drinking water and wearing a shoe.}
\label{fig:attn_visualization}
\end{figure}

\begin{table}[t]
\centering
\caption{Ablation on the strategy of unified modeling.}
\label{tab:unified}
\resizebox{0.38\textwidth}{!}{
\begin{tabular}{ccc|ccc}
\toprule
\textbf{ID} & \textbf{Params(M)} & \textbf{Acc(\%)} & \textbf{ID} & \textbf{Params(M)} & \textbf{Acc(\%)} \\
\midrule

\rowcolor{gray!30}
\multicolumn{6}{c}{\textit{Feature Alignment}} \\
1 & 5.57 & 93.49 & 2 & 5.67 & 93.67 \\

\rowcolor{gray!30}
\multicolumn{6}{c}{\textit{Topology Refinement}} \\
3 & 6.00 & 93.29 & 4 & 6.02 & 93.58 \\

\rowcolor{gray!30}
\multicolumn{6}{c}{\textit{Feature Fusion}} \\
5 & 5.80 & 93.95 & 6 & 5.97 & 93.52 \\
7 & 5.74 & 94.13 & & \\

\bottomrule
\end{tabular}}
\end{table}

\noindent \textbf{The strategy of unified modeling.} 
Based on PAN (Guided), we investigate the choice of strategies for unified modeling, as detailed in Table \ref{tab:unified}. Specifically, three directions are explored :  

(1) \textit{Feature Alignment.} We apply MSE loss to align visual token embeddings with skeletal graph embeddings — either before (ID-1) or after (ID-2) GCNs.  

(2) \textit{Topology Refinement.} We embed the learned adjacency matrices from skeletal GCNs — either per-layer (ID-3) or per-group in CTR-GC (ID-4) — and compute a weighted sum to refine the correlations in Equation \ref{topo_add}.  

(3) \textit{Feature Fusion.} We compare three choices of the fusion module in PAN-Unified: concatenation followed by $1\times 1$ convolution (ID-5), attention-based fusion (ID-6), and summation after $1\times 1$ convolution (ID-7).

Finally, we adopt ID-7 as the strategy for unified modeling and develop PAN-Unified, which achieves a significant tradeoff between accuracy and model size.

\subsection{Visualization}
In Figure \ref{fig:attn_visualization}, we provide visualizations of attention maps generated within the post calibration. The attention maps reveal that the sampled $M\times J$ tokens attend sparsely to the original visual tokens, with most attention concentrated around human joints and minimal attention on backgrounds or scenes, indicating the human-centric nature of PAN.

\section{Discussions}

\noindent \textbf{Limitations.} 
Despite its effectiveness, PAN has several limitations: (1) The sampling strategy is relatively static and lacks dynamic adaption to environments. Although the attention-based calibration mitigates this issue by refining visual token embeddings, the robustness of PAN under highly dynamic or occluded scenarios remains to be thoroughly evaluated. (2) In group activity recognition scenarios, the number of sampled tokens need to be carefully designed. Future work might explore adaptive sampling strategies or a general one for diverse group interactions.

\noindent \textbf{Future Directions.} Notably, the number of sampled tokens $M\times J$ is significantly smaller than the length of the original token sequence $\lceil \frac{H}{P}\rceil \times \lceil \frac{W}{P}\rceil$. Moreover, the basic blocks and the classification head in GCNs are lightweight and have a relatively low computational overhead. These characteristics indicates the potential of PAN for online action recognition. Another promising direction lies in unified modeling. As the visual token graphs align well with the skeletal graphs in skeleton-based methods, more sophisticated architectures for unified modeling are worth exploring. 

\section{Conclusion}
This paper proposes PAN, the first human-centric graph representation learning framework for multi-modal action recognition. PAN utilizes visual foundation models to encode RGB frames and adopts guided or even strategy to sample token embeddings. Specifically, the guided strategy leverages 2D skeletal data as guidance to sample token embeddings of patches containing human joints, while the even sampling strategy utilizes linear interpolation to gather tokens. We also propose attention-based post calibration to refine the sampled tokens, which enhances the robustness of PAN against the quality of skeletal data. 

After sampling and post calibration, the embeddings are represented as visual token graphs and processed through GCNs. Since the visual token graphs and skeletal graphs are structurally consistent and semantically aligned, the same graph modeling paradigm is applied to both, bringing the potential of fine-grained cross-modal fusion. On three widely used multimodal action recognition datasets, two variants of PAN, PAN-Ensemble and PAN-Unified, achieve state-of-the-art performance among methods adopting separate and unified modeling, respectively.
\newpage
% \input{sec/X_suppl}

% \newpage
{
    \small
    \bibliographystyle{ieeenat_fullname}
    \bibliography{main}
}

% WARNING: do not forget to delete the supplementary pages from your submission 
% \input{sec/X_suppl}

\end{document}